%% file: root.tex
\documentclass[letterpaper, 10 pt, conference]{ieeeconf}  

\usepackage{times}
\usepackage{epsfig}
\usepackage{graphicx}
\usepackage{amsmath}
\usepackage{amssymb}
\usepackage{fontawesome}

\usepackage{tikz}
\usepackage{comment}
\usepackage[utf8]{inputenc} 
\usepackage[T1]{fontenc}    
\usepackage{url}            
\usepackage{amsfonts}       
\usepackage{nicefrac}       
\usepackage{microtype}      
\usepackage[bb=boondox]{mathalfa}
\usepackage{amsmath}
\usepackage{multirow}
\usepackage{algorithmic}
\usepackage{algorithm}
 \usepackage{wrapfig}
\usepackage{rotating}
\usepackage{subfigure}
\usepackage{hyperref}

\usepackage[utf8]{inputenc} 
\usepackage[T1]{fontenc}    
\usepackage{url}            
\usepackage{booktabs}       
\usepackage{amsfonts}       
\usepackage{nicefrac}       
\usepackage{microtype}      
\usepackage{marvosym}

\usepackage{float}
\usepackage{amssymb}
\usepackage{caption}
\newcommand{\name}{MASP}
\newcommand{\hplanner}{{GM}}
\newcommand{\lplanner}{{CAE}}
\usepackage[capitalize]{cleveref}
\crefname{section}{Sec.}{Secs.}
\Crefname{section}{Section}{Sections}
\Crefname{table}{Table}{Tables}
\crefname{table}{Tab.}{Tabs.}

\IEEEoverridecommandlockouts                              

\title{\LARGE \bf 
MASP: Scalable Graph-based Planning towards Multi-Agent Navigation 
}

\author{Xinyi Yang$^{1*}$, Xinting Yang$^{1*}$, Chao Yu$^{1}$\textsuperscript{\Letter}, Jiayu Chen$^{1}$, Wenbo Ding$^{2}$, Huazhong Yang$^{1}$ and Yu Wang$^{1}$\textsuperscript{\Letter} 
\thanks{{\Letter} Corresponding Authors. \url{{yuchao,yu-wang}@tsinghua.edu.cn}}
\thanks{* Equal Contribution}
\thanks{$^{1}$Department of Electronic Engineering, Tsinghua University, Beijing, 100084, China.}%
\thanks{$^{2}$Shenzhen International Graduate School, Tsinghua University, Shenzhen, 518055, China.}%
\thanks{We acknowledge Botian Xu for providing the guidance on OmniDrones.}
\thanks{This research was supported by National Natural Science Foundation of China (No.62406159, 62325405), China Postdoctoral Science Foundation under Grant Number GZC20240830, 2024T170496.}
\thanks{More information can be found at 
\protect\url{https://sites.google.com/view/masp-ral}.
}}

\begin{document}

\maketitle
\thispagestyle{empty}
\pagestyle{empty}


\input{0_abs}
\input{1_intro}
\input{2_related}
\input{3_task}

\input{4_method}

\input{5_exp}
\input{6_con}
\bibliographystyle{IEEEtran}
\bibliography{IEEEfull}

\end{document}

%% file: 0_abs.tex
\begin{abstract}

We investigate multi-agent navigation tasks, where multiple agents need to reach initially unassigned goals in a limited time. Classical planning-based methods suffer from expensive computation overhead at each step and offer limited expressiveness for complex cooperation strategies. In contrast, reinforcement learning~(RL) has recently become a popular approach for addressing this issue. 
However, RL struggles with low data efficiency and cooperation when directly exploring (nearly) optimal policies in a large exploration space, especially with an increased number of agents~(e.g., 10+ agents) or in complex environments~(e.g., 3-D simulators).
In this paper, we propose the \emph{\underline{M}ulti-\underline{A}gent \underline{S}calable Graph-based \underline{P}lanner}~({\name}), a goal-conditioned hierarchical planner for navigation tasks with a substantial number of agents in the decentralized setting. {\color{black}{\name} employs a hierarchical framework to reduce space complexity by decomposing a large exploration space into multiple goal-conditioned subspaces, where a high-level policy assigns agents goals, and a low-level policy navigates agents toward designated goals. For agent cooperation and the adaptation to varying team sizes, we model agents and goals as graphs to better capture their relationship. The high-level policy, the Goal Matcher, leverages a graph-based Self-Encoder and Cross-Encoder to optimize goal assignment by updating the agent and the goal graphs. The low-level policy, the Coordinated Action Executor, introduces the Group Information Fusion to facilitate group division and extract agent relationships across groups, enhancing training efficiency for agent cooperation. The results demonstrate that {\name} outperforms RL and planning-based baselines in task efficiency. Compared to the planning-based competitors with a centralized goal assignment method, {\name} improves task efficiency by over 19.12\% in multi-agent particle environments with 50 agents and 27.92\% in a quadrotor 3-dimensional environment with 20 agents, achieving at least a 47.87\% enhancement across varying team sizes.}\looseness=-1



\end{abstract}


%% file: 1_intro.tex
\section{Introduction}
\label{sec:intro}
Navigation is an important task for intelligent embodied agents and is widely applied in various domains, including autonomous driving~\cite{autonomousdriving,autonomousdriving2}, logistics and transportation~\cite{logistics,logistics2}, and disaster rescue~\cite{rescue,rescue2}. This paper considers a multi-agent navigation problem where multiple agents make decisions independently and navigate simultaneously toward a set of initially unassigned goals.

Planning-based solutions have been extensively employed in multi-agent navigation tasks~\cite{frontier3,RRT_Star}. However, these methods often struggle with coordination strategies, demanding intricate hyper-parameter tuning for each specific scenario. Moreover, they can be time-consuming due to frequent re-planning at each decision step. Conversely, reinforcement learning~(RL) offers its remarkable expressiveness in multi-agent navigation tasks~\cite{maans,mage-x, inferenceMultiNavigation} with strong representation capabilities for complex strategies and low inference overhead once policies are well-trained. However, directly learning an end-to-end policy~\cite{mappo, mat} from a large exploration space results in low sample efficiency and limited cooperation, which is more severe as the number of agents or the complexity of the environment increases. {\color{black}Moreover, classical RL policies tend to overfit to fixed and trained agent numbers and often exhibit poor generalization in varying agent numbers.} Therefore, existing methods~\cite{multiagent-RL,liu2021multi} mainly focus on simple scenarios with a few fixed agent numbers.\looseness=-1

{\color{black}To solve these challenges, we follow the centralized training with decentralized execution~(CTDE) paradigm, where agents make independent decisions. 
We adopt a hierarchical framework, \emph{\underline{M}ulti-\underline{A}gent \underline{S}calable Graph-based \underline{P}lanner}~(\name). This framework consists of a high-level policy that assigns goals to agents at each global step and a low-level policy that navigates agents toward the designated goals, breaking down a large exploration space into multiple goal-conditioned subspaces. We model the agents and the goals as graphs with expandable nodes to better capture their relationships and adapt to the varying team sizes. The high-level policy, the Goal Matcher, develops a graph-based Self-Encoder and Cross-Encoder to update the agent and the goal graphs, optimizing goal assignment. The low-level policy, the Coordinated Action Executor introduces the Group Information Fusion to segment agents into groups, transform groups into subgraphs, and extract relationships between agents across subgraphs, reducing the input dimension of the network and enhancing training efficiency for agent cooperation.\looseness=-1

We compare {\name} with planning-based methods and RL-based competitors in multi-agent particle environments (MPE)~\cite{mpe} and a quadrotor environment (OmniDrones)~\cite{drone}. We remark that OmniDrones is a complex 3-dimensional (3-D) environment, further increasing the exploration space. Empirical results demonstrate that {\name} significantly outperforms its competitors in task and execution efficiency.


Our contributions are summarized as follows:
\begin{itemize}
\setlength{\parskip}{0pt} \setlength{\itemsep}{0pt plus 1pt}
\item We propose a hierarchical multi-agent navigation planner, \emph{Multi-Agent Scalable Graph-based Planner}~(\name), to enhance sample efficiency in a large exploration space.
\item We introduce the high-level policy, the Goal Matcher, using a graph-based Self-Encoder and Cross-Encoder to capture the relationships between agents and goals and finish goal assignment. In the low-level policy that navigates agents to their designated goals, we develop the Group Information Fusion to improve training efficiency and promote effective cooperation. \looseness=-1
\item Compared to the planning-based competitors using a centralized goal assignment method, {\name} increases execution efficiency by at least 3$\times$ and enhances task efficiency by over 19.12\% in MPE with 50 agents and 27.92\% in OmniDrones with 20 agents, achieving at least a 47.87\% enhancement across varying team sizes.
\end{itemize}}



\begin{figure*}[t!]
	\centering
 \vspace{2mm}
    \includegraphics[width=0.97\linewidth]{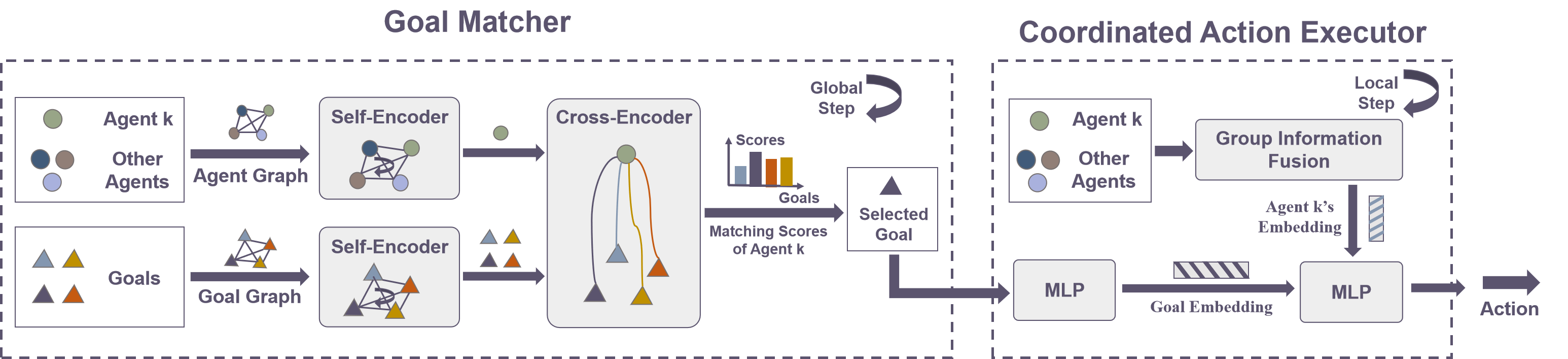}
	\centering \caption{{\color{black}Overview of the \emph{Multi-Agent Scalable Graph-based Planner}. Here, we take Agent $k$ as an example.}}
\vspace{-2mm}
\label{fig:Overview}

\end{figure*}

%% file: 2_related.tex
\section{Related Work}

\subsection{Multi-agent Navigation} 

Multi-agent navigation~\cite{maans, mage-x, save, li2020graph} is a typical cooperative task in robotics. However, finding a (nearly) optimal strategy in a large search space is difficult as the number of agents grows. \cite{li2020graph} reduces the search space by imitating an expert policy. \cite{darl1n} only considers adjacent agents to break the curse of dimensionality. However, it ignores interactions among all agents, increasing collisions. \cite{mage-x} introduces a hierarchical approach with centralized goal allocation and a subgraph construction for agent cooperation. Nonetheless, \cite{mage-x} can not adapt to varying agent numbers without an agent-invariant design. This paper proposes a graph-based hierarchical framework that adapts to dynamic agent numbers in MPE and a 3-D quadrotor simulator, Omnidrones.\looseness=-1


\subsection{Goal Assignment}
In multi-agent navigation, assigning suitable targets to agents enhances efficiency~\cite{kuhn1955hungarian, turpin2014goal, ma2017overview}. 
The Hungarian algorithm~\cite{kuhn1955hungarian} uses a weighted bipartite graph to find a matching with maximum weight. However, it requires equal numbers of agents and goals, limiting its use in scenarios with dynamic agent numbers. \cite{turpin2014goal} allocates goals to minimize the longest path of the trajectories. However, the high time expenditure makes it difficult to extend to large-scale multi-agent settings. In \cite{wagner2011m}, the targets are only reassigned when a collision is about to occur, effectively reducing the computational costs. Nevertheless, this method ignores the collaboration between agents. 
This work proposes an RL-based goal assignment method that can extend to scenarios with large and varying agent numbers and ensure effective cooperation.\looseness =-1



\subsection{Goal-conditioned HRL}
Goal-conditioned hierarchical reinforcement learning (HRL) is effective across various tasks~\cite{ji2022hierarchical, nasiriany2022augmenting}, typically using a high-level policy to select subgoals and a low-level policy to predicts environmental actions to achieve these subgoals.
{\color{black}However, existing approaches~\cite{takubo2022hierarchical, pope2021hierarchical, hafner2022deep} struggle with scenarios with multiple initially unassigned targets, failing to establish connections between subgoals and multiple final targets. \cite{mage-x} and \cite{htamp} adopt a hierarchical framework in multi-agent navigation, where the high-level policy assigns each agent a target, and the low-level policy yields the environmental actions based on the assigned target.
Inspired by \cite{mage-x,htamp}, we develop a goal assignment module as the high-level policy to assign agents with final targets.}

%% file: 3_task.tex
\section{Preliminary}
{\color{black}
We formulate multi-agent navigation tasks as goal-conditioned Markov Decision Processes~(MDPs) $M = ⟨\mathcal{N}, \mathcal{G}, S, A, P, R, \gamma⟩$. $\mathcal{N}\equiv\{1,...,N\}$ is the set of $N$ agents. $\mathcal{G}$ is a goal set. $S$ is the joint state space. $A$ is the joint action space. $P$ is the transition function. $R$ is the reward function. $\gamma$ is the discount factor.
In our task, we establish a hierarchical framework with a high-level policy $\pi_{\theta_h}^{h}(g|s)$ and a low-level policy $\pi_{\theta_l}^{l}(a|s,g)$ parameterized by two neural networks with parameters $\theta_h$ and $\theta_l$ respectively. Each agent's state space includes the locations and the velocities of all agents and the locations of goals. The high-level policy aims to maximize the accumulative reward for goal assignment and generates a high-level action, i.e. a goal $g_t\sim\pi_{\theta_h}^{h}(g|s)\in \mathcal{G}$ every global step, i.e., 3 timesteps. The low-level policy receives these goals, seeks to maximize the accumulative reward for goal achievement, and performs a local action $a_t\sim\pi_{\theta_l}^{l}(a|s,g)\in A$ at every timestep.}



%% file: 4_method.tex
\section{Methodology}
\subsection{Overview}
To solve the issues of low sample efficiency in navigation tasks with substantial numbers of agents, we adopt a hierarchical framework, consisting of a high-level policy, the Goal Matcher~({\hplanner}), that assigns agents with goals, and a low-level policy, the Coordinated Action Executor~({\lplanner}), that encourages agents to navigate towards the designated goals. This divides a large exploration space into multiple goal-conditioned subspaces, reducing space complexity and speeding up training. 
{\color{black}Moreover, we model the agents and goals as graphs with expandable nodes to facilitate cooperation and adapt to varying team sizes.} As depicted in \cref{fig:Overview}, we introduce this overall framework, the \emph{Multi-Agent Scalable Graph-based Planner}. \looseness=-1 


Take agent $k$ as an example. We first consider two graphs: the agent graph representing agent information and the goal graph representing goal information.
{\hplanner} takes in these graphs to capture the relationship among agents and goals, and yields the most appropriate goal for agent $k$ at each global step. 
In {\lplanner}, agent $k$ extracts the designated goal’s features as a goal embedding. 
{\color{black}Meanwhile, to reduce the input dimension of the network and improve training efficiency for agent cooperation in {\lplanner}, we develop the Group Information Fusion to segment the agents into groups, transform groups into subgraphs, and capture the relationships among agents across subgraphs. The Group Information Fusion provides an updated feature of agent $k$ across all subgraphs, which, along with the goal embedding, passes through a multi-layer perception~(MLP) layer to endow the team representation with strong goal guidance. Finally, agent $k$ takes environmental actions to navigate towards the goal based on the extracted representation.\looseness=-1

We adopt the centralized training with decentralized execution~(CTDE) paradigm, where agents make independent decisions. We train {\hplanner} and {\lplanner} simultaneously by using multi-agent proximal policy optimization (MAPPO)~\cite{mappo}, a multi-agent variant of proximal policy optimization (PPO)~\cite{ppo}.}

\subsection{Goal Matcher}
Goal assignment is a long-studied NP-hard maximum matching problem~\cite{turpin2014goal,gerkey2004formal}. The Hungarian algorithm~\cite{Hungarian} is a classical combinatorial optimization algorithm that utilizes the bipartite graph matching to solve the assignment problem. 
{\color{black}However, it assigns goals centrally, requiring equal numbers of agents and goals. With varying agent numbers, we need to first assign a part of the goals, and then assign the remaining goals, causing inefficiency. To solve this problem, we adopt a decentralized approach where each agent selects a goal independently at each global step, allowing flexibility in goal selection and adjustment. Additionally, we leverage graphs with expandable nodes to represent agents and goals respectively, handling the issue of varying agent numbers. 



However, this decentralized setting potentially leads to multiple agents pursuing the same goal—something impossible under centralized decision-making.} For better goal assignment, we introduce an RL-based high-level policy, the Goal Matcher. In {\hplanner}, agent $k$ first receives two fully connected graphs: the agent graph, ${G_{A}}(V,E)$, and the goal graph, ${G_{T}}(V,E)$. $V$ is the node set representing the position information of agents or goals, and $E$ is the edge set representing the connection between agents or goals. {\color{black}The edge features in $E$ are all initially set to $1$. Afterward, we leverage the Self-Encoder to perceive the spatial relationship between the agents or goals and update the node and the edge features.} We then use the Cross-Encoder to extract the relationship between agent $k$ and the goals and calculate the matching score of agent $k$. 
Finally, agent $k$ chooses the goal with the highest matching score at each global step.


{\color{black}The reward for agent $k$, $R_{GM}^k$, encourages agents to select different goals with minimal total distance, which is formulated as follows:


\begin{equation}
        R_{GM}^k = {\begin{array}{l}\left\{\begin{array}{l}
        \begin{array}{l}0,\qquad\qquad\qquad g_{m}^k==g_{h}^k;\\-(1+\frac{N_{repeat}}{N}),\;g_{m}^k\neq g_{h}^k and\;N_{repeat}>0;\\-(1-\frac{D_{h}}{D_{m}}),\qquad g_{m}^k\neq g_{h}^k and\; N_{repeat}=0.\end{array}\end{array} \right.
        \\
        \end{array}}
\end{equation}
Here, $N_{repeat}$ signifies the number of other agents whose predicted goal is the same as agent $k$. We use $N_{repeat}$ to promote agents to select different goals. $g_{m}^k$ represents the predicted goal from {\hplanner} for agent $k$, and $g_{h}^k$ denotes the goal assigned by the Hungarian algorithm. $D_{m}$ and $D_{h}$ denote the total distance from the predicted and Hungarian-assigned goals to the agents, respectively.
If our method has a lower distance cost,
$R_{GM}^k$ becomes positive, incentivizing the Goal-Matcher to learn an better goal assignment strategy. }




\subsubsection{Self-Encoder} {\color{black}We use the Self-Encoder to update ${G_{A}}$ and ${G_{T}}$ respectively. It captures the relationships between any two nodes and updates the edges and nodes in the graph. Inspired by the attention mechanism~\cite{attention}, we first compute a matrix, $S_{inter}\in\mathbb{R}^{N\times N}$, as the updated edge features.


\begin{equation}
S_{inter} = Softmax\left(\left(W_{query} X^T\right){\left(W_{key} X^T\right)}^T\right),
\end{equation}
where $X\in\mathbb{R}^{N\times L}$ signifies $N$ node features, each with $L=32$ dimensions, and $W\in\mathbb{R}^{N\times L}$ denotes the linear projections of $X$. }

Each node then updates its feature by embedding it with a weighted sum of its neighbors' features via an MLP layer.\looseness=-1

{\color{black}\begin{equation}
    \begin{aligned}
        Y = X
        +MLP\left(Concat\left(X,  \left(W_{value} {X}^T\right) {S_{inter}}^T\right)\right),
    \end{aligned}
        \end{equation}}


\subsubsection{Cross-Encoder}{\color{black}The Cross-Encoder takes in the updated node feature for agent $k$, ${{Y_A}^k}\in\mathbb{R}^{1\times L}$, and all updated node features for goals, ${Y_T}\in\mathbb{R}^{N\times L}$, where $L$ is 32 in our work.} 
It captures the position correlations between agent $k$ and goals and calculates matching scores between them, ${S_{intra}^k}$, using the node features and L2 distance, $D$, via the attention mechanism~\cite{attention}. The calculation is formulated as follows:
\begin{equation}
   {S_{intra}^k = Softmax\left( MLP\left(Concat\left(\tilde{{Y_A}^k},\tilde{Y_T}
    , D\right)\right)\right)},
\end{equation}
where $W$ is the linear projections of ${{Y_A}^k}$ or ${Y_T}$. $\tilde{{Y_A}^k}$ is $W_{query}{{Y_A}^k}^T$ and $\tilde{Y_T}$ is $W_{key} {Y_T}^T$. 







    
\begin{figure}[ht]
	\centering
    \includegraphics[width=0.97\linewidth]{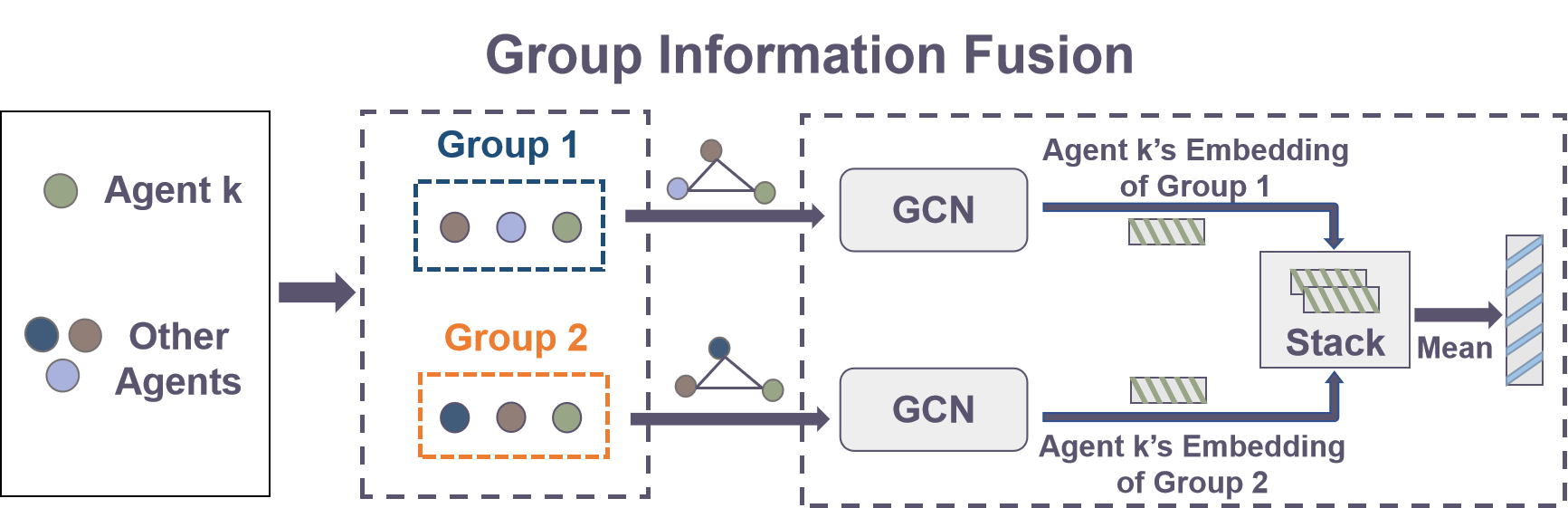}
	\centering \caption{{\color{black}The workflow of the Group Information Fusion in the \emph{Coordinated Action Executor}. Take Agent $k$ as an example.}}
\label{fig:division}
\vspace{-5mm}
\end{figure}

\subsection{Coordinated Action Executor}

We introduce the Coordinated Action Executor as the low-level policy to encourage agents to reach the assigned goals. {\color{black}Firstly, to reduce the input dimension of the network and enhance the training efficiency for agent cooperation, we develop the Group Information Fusion to divide the agents into groups and extract the relationships between agents across groups. 
\cref{fig:division} illustrates the workflow of the Group Information Fusion for agent $k$. For the group division in the Group Information Fusion, agent $k$ is required to participate in every group to maintain connections with other agents. Each group consists of $N_g=3$ agents, so we first divide the remaining agents, excluding agent $k$, into groups, each consisting of $N_g-1$ agents. If the remaining agents can not be evenly divided, we randomly select one agent to join more than one group to ensure group formation. Finally, we add agent $k$ to each group to complete the group division process.
We transform each group into a fully connected subgraph, where the nodes contain the state information of the agents, and the edges are set to 1. The Group Information Fusion utilizes graph convolutional networks~(GCN) to perceive the interaction between agent $k$ and other agents in each group and update the node feature. Subsequently, to obtain the updated feature of agent $k$ that infuses the relationship between agent $k$ and all other agents, we aggregate the updated node feature of agent $k$ across all groups via a mean operation over the group dim. \looseness=-1

Meanwhile, we extract the designated goal embedding for agent $k$ via an MLP layer. 
Receiving the goal embedding and the updated feature of agent $k$, we apply another MLP layer to perceive the correlation between the designated goal and agent $k$.
Finally, based on the extracted correlation, we produce the environmental action for agent $k$ to navigate toward the goal.}

The reward, $R_{CAE}^k$, for agent $k$ in the {\lplanner} promotes goal-directed navigation while minimizing collisions. $R_{CAE}^k$ is a linear combination of the complete bonus, $R_b$, the distance penalty, $R_d$, and the collision penalty, $R_c$:

\begin{equation}
    R_{CAE}^k = \alpha R_{b}^k+\beta R_{d}^k+\gamma R_{c}^k,
\end{equation}
where $\alpha$, $\beta$ and $\gamma$ are the coefficients of $R_{b}^k$, $R_{d}^k$ and $R_{c}^k$, respectively.

%% file: 5_exp.tex
\section{Experiments}



\subsection{Testbeds}
To evaluate the effectiveness of our approach in large exploration spaces, we opt for two environments: MPE~\cite{mpe} and OmniDrones~\cite{drone}, with a substantial number of agents.
\subsubsection{MPE}
MPE is a classical 2-dimensional environment. The collision between agents causes disruptive bounce-off, reducing navigation efficiency. The available environmental actions of the agents are discrete, including \emph{Up}, \emph{Down}, \emph{Left}, and \emph{Right}. The experiment is conducted with randomized spawn locations for $N$ agents, $N$ landmarks and $B$ obstacles.
The agents, the obstacles, and the landmarks are circular with a radius of 0.1$m$, 0.1$m$, and 0.05$m$, respectively.
We consider $N\in\{5, 20, 50\}$ and $B\in\{5, 10, 30\}$ on the maps of 4$m^2$, 64$m^2$, and 400$m^2$, with the horizons of the environmental steps of 18, 45, and 90, respectively. 


\subsubsection{OmniDrones}
{\color{black}We also conduct experiments in OmniDrones, an efficient and flexible 3-D simulator for drone control, which further increases the exploration space.} The continuous action space of each drone is the throttle for each motor. Collisions directly lead to crashes and task failures, increasing task difficulty. The drones are hummingbirds with 0.17$m$ arms and 4 motors. The landmarks are virtual balls of 0.05$m$ radius.
We conduct experiments with $N\in\{5, 20\}$ drones, where the spawn locations for both drones and landmarks are randomly distributed on the maps sized at 36$m^2$ and 256$m^2$, respectively. Besides, the horizons of the environmental steps are 300 for 5 drones and 500 for 20 drones.\looseness=-1


\begin{figure*}[ht!]
	\centering
 \vspace{2mm}
    \includegraphics[width=0.7\linewidth]{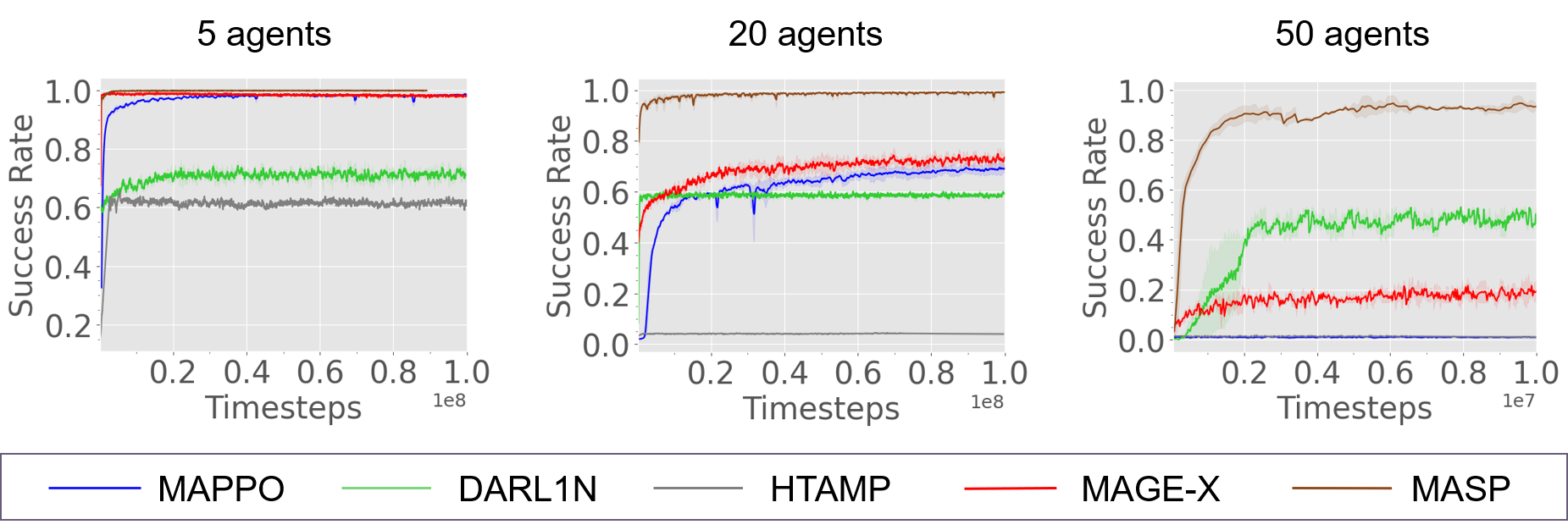}
  
	\centering \caption{Comparison between {\name} and other baselines in MPE with $N=5$, $20$, $50$.}
 \vspace{-2mm}
\label{fig:mpe}

\end{figure*}
\begin{figure}[ht!]
\vspace{-0mm}
	\centering
    \includegraphics[width=0.9\linewidth]{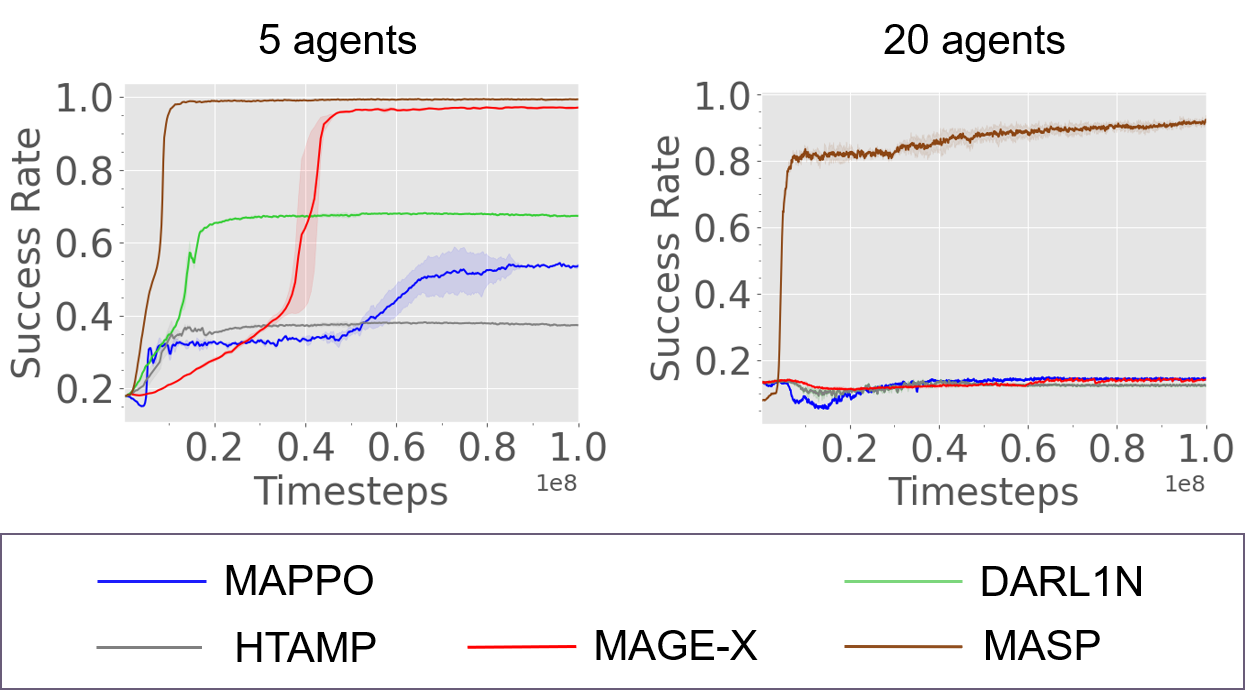}
  
	\centering \caption{\raggedright{Comparison between {\name} and other baselines in Omnidrones with $N=5$, $20$.}}
\label{fig:omnidrones}
 \vspace{-5mm}
\end{figure}

\subsection{Implementation Details}
Each RL training is performed over 3 random seeds for a fair comparison. {\color{black}Each evaluation score is expressed in the format of "mean (standard deviation)", averaged over a total of 300 testing episodes, i.e., 100 episodes per random seed.}

\subsection{Evaluation Metrics}
We consider 3 statistical metrics to capture different characteristics of a particular navigation strategy. 

\begin{itemize}
\setlength{\parskip}{0pt} \setlength{\itemsep}{0pt plus 1pt}
    {\color{black}\item \textbf{Success Rate~(SR): }This metric measures the average ratio of landmarks reached by the agents to the total landmarks per episode. \looseness=-1
    \item \textbf{Steps: }This metric represents the average timesteps required to achieve a target \emph{Success Rate} per episode.
    \item \textbf{Collision Rate~(CR): }This metric denotes the average ratio of collision occurrences to the total number of environmental steps per episode.}
    
\end{itemize}
We remark that we consider \emph{Steps} as the primary metric for measuring task efficiency. In the tables, the backslash in Steps denotes the agents can not reach the target 100\% \emph{Success Rate} in any episode.

\subsection{Baselines}
We compare {\name} with three representative planning-based approaches (ORCA, RRT*, Voronoi) and four prominent RL-based solutions (MAPPO, DARL1N, HTAMP, MAGE-X). In planning-based approaches, we utilize the Hungarian algorithm represented as `(H)' for goal assignment.

\begin{itemize}
\setlength{\parskip}{0pt} \setlength{\itemsep}{0pt plus 1pt}
  \item \textbf{ORCA}~\cite{orca}: ORCA is an obstacle avoidance algorithm that excels in multi-agent scenarios. It predicts the movements of surrounding obstacles and other agents, and then infers a collision-free velocity for each agent.  
  \item \textbf{RRT*}~\cite{RRT_Star}: RRT* is a sample-based planning algorithm that builds upon the RRT algorithm~\cite{RRT}. It first finds a feasible path by sampling points in the space and then iteratively refines the path to achieve an asymptotically optimal solution.
 \item \textbf{Voronoi}~\cite{Voronoi}: A Voronoi diagram comprises a set of continuous polygons, with a vertical bisector of lines connecting two adjacent points. By partitioning the map into Voronoi units, a path can be planned from the starting point to the destination at a safe distance.
 \item \textbf{MAPPO}~\cite{mappo}: MAPPO is a straightforward extension of PPO in the multi-agent setting, where each agent is equipped with a policy with a shared set of parameters. We update this shared policy based on the aggregated trajectories from all agents. We additionally apply the attention mechanism to enhance the model's performance.
 \item \textbf{DARL1N}~\cite{darl1n}: DARL1N is under an independent decision-making setting for large-scale agent scenarios. It breaks the curse of dimensionality by restricting the agent interactions to one-hop neighborhoods. 
 \item \textbf{HTAMP}~\cite{htamp}: This hierarchical method for task and motion planning via reinforcement learning integrates high-level task generation with low-level action execution.\looseness=-1
 \item \textbf{MAGE-X}~\cite{mage-x}: This is a hierarchical approach to multi-agent navigation tasks. It first centrally allocates target goals to the agents at the beginning of the episode and then utilizes GNN to construct a subgraph only with important neighbors for higher cooperation.

\end{itemize}

\input{tables/mpe}

\input{tables/omnidrones}
\subsection{Main Results}
\subsubsection{Training with a Fixed Team Size}
\textbf{\\MPE: }We present the training curves in Fig.~\ref{fig:mpe} and the evaluation performance in \cref{tab: mpe}. {\name} outperforms all methods with the least training data, with its advantage in \emph{Steps} and \emph{Success Rate} becoming more evident as agent numbers increase. {\name} presents the lowest \emph{Collision Rate}, reflecting strong agent cooperation.
Especially in scenarios with 50 agents, {\name} achieves nearly 100\% \emph{Success Rate}, while other RL baselines fail to complete the task in any episode.
Among RL baselines, DARL1N has a 49\% \emph{Success Rate} at $N=50$, while it shows the worst performance at $N=5$. This implies that when the scenario only has a few agents, it may lead to the absence of one-hop neighbors, thereby influencing the agents' decisions.
MAGE-X has a 23\% \emph{Success Rate} at $N=50$, suggesting its MLP-based goal assignment encounters challenges in preventing collisions between a large number of agents.
MAPPO has only a 0.01\% \emph{Success Rate} at $N=50$, showing difficulty in extracting agent correlations and identifying goals. 
(H)MAPPO has a comparable performance with a fixed agent number and requires 8.52\% more \emph{Steps} than {\name} at $N=50$. 
Although HTAMP has a hierarchical framework, its MLP backbone fails to perceive the relationship between agents and targets, exhibiting inferior performance.

{\color{black}Regarding planning-based methods, (H)RRT* and (H)Voronoi shows a comparable performance. This suggests that (H)RRT* omits complex constraints, suitable for scenarios with obstacles. (H)Voronoi partitions the map to promise coordination. However, they still require around 19.12\% more \emph{Steps} than {\name} at $N=50$. (H)ORCA performs worse, needing at least 21.76\% more \emph{Steps} than {\name}, highlighting their difficulty in finding shorter navigational paths.}


\textbf{OmniDrones:}
We report the training performance in Fig.~\ref{fig:omnidrones} and the evaluation results in \cref{tab: omnidrones}. {\name} is superior to all competitors, achieving a 100\% \emph{Success Rate} at $N=5$ and 96\% at $N=20$ in this complex 3-D environment. As for the \emph{Collision Rate}, {\name} outperforms other baselines due to high agent cooperation. 
Most RL baselines, except MAGE-X and (H)MAPPO, achieve only around 50\% \emph{Success Rate} at $N=5$, failing to address the navigation problem in the complex search space. However, MAGE-X drops to a 15\% \emph{Success Rate} at $N=20$. This reveals that its MLP-based goal assignment struggles with large and complex environments.\looseness=-1

{\color{black}In contrast to RL baselines, most planning-based baselines achieve over 90\% \emph{Success Rate}, primarily owing to the goal assignment algorithm without duplication. Thus, we focus more on \emph{Steps} for comparison. (H)Voronoi is the best planning-based competitor at $N=5$, while all the planning-based baselines show comparable performance at $N=20$. This indicates that although (H)Voronoi's partitions excel in scenarios with a few agents, they are difficult to figure out a set of coordinated navigation paths in complex 3-D environments with increased agent numbers. 
The planning-based baselines requires at least 27.92\% more \emph{Steps} than {\name} at $N=20$.}\looseness=-1

\input{tables/mpe_varying}

\input{tables/omnidrones_varying}
\subsubsection{Varying Team Sizes within an Episode}
{\color{black}Due to unstable communication or agent loss, we further consider scenarios where the team size decreases during an episode. With more goals than agents, some agents may need to first reach some goals and then adjust their plans to navigate towards the remaining goals. The \emph{Success Rate} is defined as  the ratio of the landmarks reached during the navigation to the total number of landmarks. 
We use "$N_1\Rightarrow N_2$" to denote that an episode starts with $N_1$ agents and switches to $N_2$ after one-third of the total timesteps. The number of landmarks is $N_1$.
Since {\name} is trained with a fixed team size, this setup presents a zero-shot generalization challenge. We use the {\name} model trained with $N=N_1$.
For the Hungarian algorithm which centrally assign agents goals, we randomly select a subset of unreached goals matching the agent count at each global step.} 
As RL baselines lack generalization in both network architecture and final performance, we compare {\name} only with planning-based methods.


 
\textbf{MPE: }As shown in \cref{tab: mpe_varying}, compared to the setting with a fixed number of agents, the advantage of {\name} becomes more apparent. {\name} consumes around 47.87\% fewer \emph{Steps} than (H)RRT* and (H)Voronoi in the $50\Rightarrow 20$ scenario. This indicates that using the Hungarian algorithm for goal assignment is more challenging for varying team sizes.

\textbf{OmniDrones: }\cref{tab:omnidrones_varying} presents evaluation results for zero-shot generalization in OmniDrones. The planning-based baselines fail to reach a 100\% \emph{Success Rate} in any episode.
This challenge stems from the absence of certain agents, therefore the remaining agents need to dynamically adjust their plans to reach more than one goal.
This is particularly difficult in such a complex 3-D environment. Conversely, {\name} maintains an average \emph{Success Rate} of 94\% due to its flexible and adaptive strategy.\looseness=-1

\input{tables/time_table}

\input{tables/ab}
\subsection{Execution Efficiency}
To evaluate the execution efficiency, we compute the per-step time of baselines and {\name} in MPE with $50$ agents, excluding environmental execution time. In Tab.~\ref{tab: time_table}, {\name} is at least 3$\times$ more efficient than planning-based competitors. Especially compared to the best planning-based competitor, (H)RRT*, {\name}'s computation time is 32$\times$ faster than it. While {\name} has a comparable execution efficiency to RL-based methods, it outperforms them in task performance. \looseness=-1

{\color{black}\subsection{Sensitivity Analysis}
\subsubsection{Group Size}
As shown in Tab.~\ref{tab: groupsize}, we perform a sensitivity analysis of group size for group division in {\lplanner} in MPE with 20 agents. When the group size is too small, it hinders capturing the correlations between agents. A larger group size improves the performance but increases network input dimensionality, depreciating the training efficiency. The performance improvement of {\name} is negligible when the group size exceeds 3. Thus, we select a group size of 3 to balance the final performance and the training efficiency.

\subsubsection{Global Steps}
We report the evaluation performance in Tab.~\ref{tab: globalsteps} to assess the sensitivity to global steps in MPE with 20 agents. Fewer global steps lead to more frequent goal assignment adjustments, improving the final performance. However, it also increases execution time. The results indicate that the performance remains comparable when the global steps are fewer than 3. To balance the final performance and the execution time, we set the global steps to 3.}

\input{tables/groupsize}

\input{tables/globalsteps}
\subsection{Ablation Studies}

To illustrate the effectiveness of each component of {\name}, we consider 3 variants of our method in MPE with 20 agents:
\begin{itemize}
\setlength{\parskip}{0pt} \setlength{\itemsep}{0pt plus 1pt}
 \item \textbf{{\name} w. RG: }We substitute {\hplanner} with random sampling without replacement to assign each agent a random target goal. \looseness=-1
 \item \textbf{{\hplanner} w.o. Graph: }We consider the MLP layer as an alternative to {\hplanner} for assigning target goals to each agent at each global step. The MLP layer takes in the concatenated position information of all agents and goals for agent $k$, with the position information of agent $k$ listed first.
 \item \textbf{{\lplanner} w.o. Graph: }We use the MLP layer as an alternative to {\lplanner} to capture the correlation between agents and goals. The input for agent $k$ comprises the position of all agents and the assigned goal for agent $k$.
\end{itemize}

\begin{wrapfigure}{l}{4.5cm}
	\raggedleft
  \vspace{-1mm}
    \includegraphics[width=0.265\textwidth]{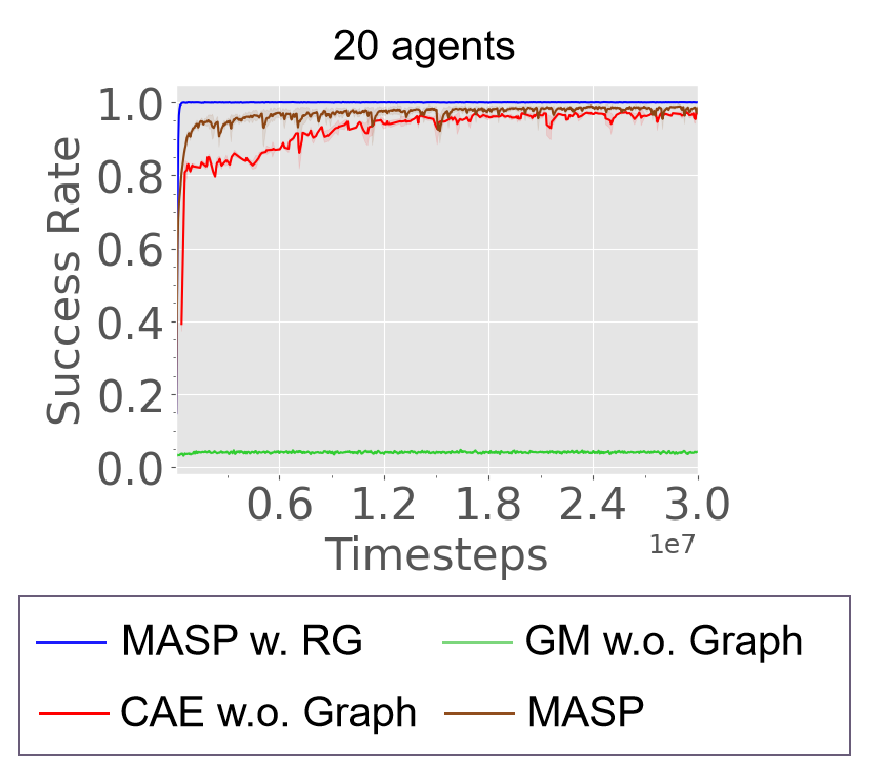}
	\caption{Ablation study on {\name} in MPE with $20$ agents.}
 \vspace{-4mm}
\label{fig: ab}
\end{wrapfigure}

To demonstrate the performance of {\name} and its variants, we report the training curves in Fig.~\ref{fig: ab} and the evaluation results in \cref{tab: ab}. 
{\name} outperforms its counterparts with a 100\% \emph{Success Rate} and the fewest \emph{Steps}. 
\emph{{\hplanner} w.o. Graph} degrades the most, indicating difficulties in perceiving the correlations between agents and target goals without graph-based Self-Encoder and Cross-Encoder. 
{\color{black}\emph{{{\name} w. RG}} only trains the low-level policy, leading to faster training convergence. During evaluation, it consistently assigns different goals to agents, achieving a 100\% \emph{Success Rate}. However, it requires 32.02\% more \emph{Steps} due to the lack of an appropriate goal assignment.} \emph{{\lplanner} w.o. Graph} is slightly inferior to {\name} with slower training convergence and 13.59\% more \emph{Steps} in the evaluation. This suggests that the GNN in {\lplanner} better captures the relationships between agents and assigned goals than the MLP layer.\looseness=-1

\subsection{Strategies Analysis}


\begin{figure}[ht]
\captionsetup{justification=centering}
\vspace{-2mm}
 \centering
    \subfigure[\label{fig:case1}Learned strategy of DARL1N.]
        {\centering
        {\includegraphics[height=2.8cm]{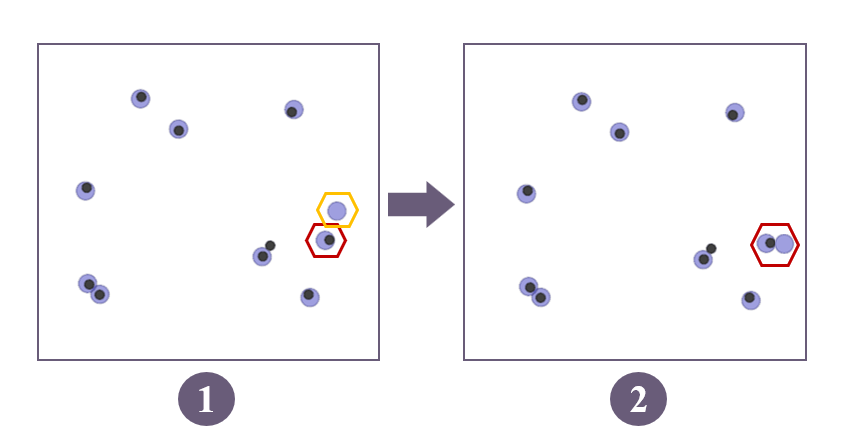}
            }
    }
    
    \subfigure[\label{fig:case2}Learned strategy of {\name}.]
        {\centering
        {\includegraphics[height=2.8cm]{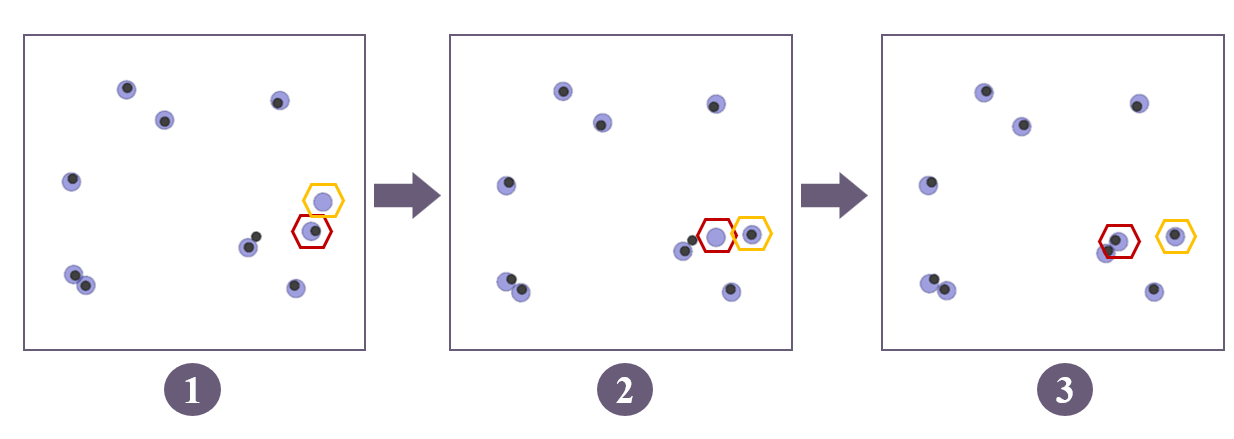}
    	}
    }
     \vspace{-2mm}
    \caption{\raggedright {The comparison in learned strategies of DARL1N and {\name} in MPE.  }}
    \label{fig:case}
    \vspace{-3mm}
\end{figure} 

As depicted in Fig.~\ref{fig:case}, we compare the learned strategies of DARL1N and {\name} to showcase the high cooperation among {\name}'s agents. Specifically, agents making independent decisions potentially lead to situations where the agent in the red pentagon and the agent in the yellow pentagon select the same goal. In Fig.~\ref{fig:case1}, these two DARL1N agents fail to adjust their strategies and finally stay at the same target. In contrast, Fig.~\ref{fig:case2} illustrates that when the red {\name} agent notices the yellow {\name} agent moving toward the same goal, it quickly adjusts its strategy to navigate toward an unexplored goal.  
This cooperation allows {\name} agents to occupy different goals, achieving a 100\% \emph{Success Rate}.\looseness=-1


%% file: tables/mpe.tex
\begin{table*}
{\color{black}\centering
\vspace{2mm}
\footnotesize
\scalebox{0.98}{
\setlength\tabcolsep{1.5mm}{\begin{tabular}{crccc|ccccc|c} 
\toprule
Agents                                                                                        & Metrics                  &(H)ORCA  &(H)RRT* &(H)Voronoi &(H)MAPPO & MAPPO &DARL1N&HTAMP&MAGE-X & {\name}  \\ 
\midrule

\multirow{3}{*}{N =5}& \textit{Steps} $\downarrow$ 
                    &16.38\scriptsize{(0.53)} &7.58\scriptsize{(1.49)}& 8.86\scriptsize{(1.27)} 
                 &      8.04\scriptsize{(1.53)}          &8.69\scriptsize{(1.24)}&
                 $\backslash$&$\backslash$& 14.75\scriptsize{(0.60)} & \textbf{7.54\scriptsize{(1.37)}} \\ 
\cmidrule{2-11}
                     & \textit{SR} $\uparrow$    &0.72\scriptsize{(0.01)}&  0.99\scriptsize{(0.01)}&    0.98\scriptsize{(0.01)} &\textbf{1.00\scriptsize{(0.00) }}
                   &\textbf{1.00\scriptsize{(0.00) }}&0.74\scriptsize{(0.02) } &0.65\scriptsize{(0.05)}&   \textbf{1.00\scriptsize{(0.00) }}&    \textbf{1.00\scriptsize{(0.00) }}    \\ 
\cmidrule{2-11}
                     & \textit{CR} $\downarrow$    & 0.57\scriptsize{(0.03)}    &3.41\scriptsize{(0.14)}
                   &3.44\scriptsize{(0.25)}
                    &0.42\scriptsize{(0.02)}& 1.32\scriptsize{(0.38)}&2.80\scriptsize{(0.53)} &1.63\scriptsize{(0.43)}& 0.45\scriptsize{(0.02)} &  \textbf{0.47\scriptsize{(0.01)}}    \\ 
\midrule
\multirow{3}{*}{N =20}& \textit{Steps} $\downarrow$   
                  &34.28\scriptsize{(3.22)} &26.86\scriptsize{(2.27)} &28.72\scriptsize{(2.68)}  & 27.43\scriptsize{(2.49)}&$\backslash$& $\backslash$&$\backslash$ &$\backslash$ & \textbf{25.61\scriptsize{(2.10)}} \\ 
\cmidrule{2-11}
                        & \textit{SR} $\uparrow$    &0.95\scriptsize{(0.03)}&  0.99\scriptsize{(0.01)}&    \textbf{1.00\scriptsize{(0.00)}}&
                          \textbf{1.00\scriptsize{(0.00) }}& 0.71\scriptsize{(0.02) }&0.62\scriptsize{(0.03) }&  0.05\scriptsize{(0.01) } &0.75\scriptsize{(0.02) }&    \textbf{1.00\scriptsize{(0.00) }}    \\ 
\cmidrule{2-11}
                         & \textit{CR} $\downarrow$    &0.49\scriptsize{(0.04)}& 1.69\scriptsize{(0.05)} &     
                   1.80\scriptsize{(0.03)} &0.75\scriptsize{(0.05)}&2.53\scriptsize{(1.02)} &3.22\scriptsize{(1.23)} &1.54\scriptsize{(0.05)}&1.78\scriptsize{(0.14)}  & \textbf{0.44\scriptsize{(0.03)} }  \\ 
\midrule
\multirow{3}{*}{N = 50} & \textit{Steps} $\downarrow$   
&60.66\scriptsize{(4.76)}&58.68\scriptsize{(3.89)} &57.88\scriptsize{(5.43)} & 51.88\scriptsize{(4.06)}&$\backslash$&$\backslash$ & $\backslash$&$\backslash$ &\textbf{47.46\scriptsize{(5.93)}}\\ 
\cmidrule{2-11}
                        & \textit{SR} $\uparrow$    &0.97\scriptsize{(0.01)}&\textbf{0.99\scriptsize{(0.01)}}&\textbf{0.99\scriptsize{(0.01)}}& \textbf{0.99\scriptsize{(0.01)}}& 0.01\scriptsize{(0.01)}&0.49\scriptsize{(0.02)}&0.02\scriptsize{(0.01)}
                      &  0.23\scriptsize{(0.01)}& \textbf{0.99\scriptsize{(0.01)} }\\ 
\cmidrule{2-11}
                        & \textit{CR} $\downarrow$   &0.30\scriptsize{(0.02)}&  1.39\scriptsize{(0.14)} &1.31\scriptsize{(0.42)} 
                    &0.42\scriptsize{(0.75)}& 0.78\scriptsize{(0.26)}&1.14\scriptsize{(0.34)} &1.35\scriptsize{(0.21)}& 1.69\scriptsize{(0.37)} & \textbf{0.23\scriptsize{(0.01)}}\\ 

\bottomrule
\end{tabular}}}}
\caption{Performance of {\name}, planning-based baselines and RL-based baselines with $N=5,20,50$ agents in MPE.}
\label{tab: mpe}
\end{table*}

%% file: tables/omnidrones.tex
\begin{table*}
\centering
\footnotesize
{\color{black}
\scalebox{0.95}{
\setlength\tabcolsep{1.5mm}{\begin{tabular}{crccc|ccccc|c} 
\toprule
Agents                                                                                        & Metrics                  &(H)ORCA  &(H)RRT* &(H)Voronoi &(H)MAPPO  & MAPPO &DARL1N&HTAMP&MAGE-X & {\name}  \\ 
\midrule

\multirow{3}{*}{N =5}& \textit{Steps} $\downarrow$   
                  &260.10\scriptsize{(10.02)} &236.10\scriptsize{(9.47)}& 173.89\scriptsize{(12.91)} &185.34\scriptsize{(7.14)} &$\backslash$&$\backslash$ & $\backslash$ &231.26\scriptsize{(19.32)} & \textbf{138.35\scriptsize{(13.57)}} \\ 
\cmidrule{2-11}
                     & \textit{SR} $\uparrow$    &0.98\scriptsize{(0.03)}&  0.97\scriptsize{(0.01)}&    0.99\scriptsize{(0.02)} &  \textbf{1.00\scriptsize{(0.00) }}& 0.50\scriptsize{(0.09) }& 0.68\scriptsize{(0.02) }& 0.32\scriptsize{(0.08) } & 0.99\scriptsize{(0.01) }&    \textbf{1.00\scriptsize{(0.00) }}    \\ 
\cmidrule{2-11}
                     & \textit{CR} $\downarrow$    &\textbf{0.00\scriptsize{(0.01) }}& 0.01\scriptsize{(0.01) } &0.02\scriptsize{(0.01) } &0.01\scriptsize{(0.01) }& 0.01\scriptsize{(0.01) }&0.01\scriptsize{(0.01) }& 0.01\scriptsize{(0.01) } & 0.01\scriptsize{(0.01) }&    \textbf{0.00\scriptsize{(0.00) }}       \\ 
\midrule
\multirow{3}{*}{N =20}& \textit{Steps} $\downarrow$   
                  &449.10\scriptsize{(14.86)} &438.23\scriptsize{(11.02)}& 437.43\scriptsize{(14.24)}&335.74\scriptsize{(13.07)}
                  &$\backslash$&$\backslash$
                  &$\backslash$ &
                  $\backslash$ & \textbf{315.31\scriptsize{(14.52)}} \\ 
\cmidrule{2-11}
                        & \textit{SR} $\uparrow$    &\textbf{0.97\scriptsize{(0.03)}}&  \textbf{0.97\scriptsize{(0.01)}}&    0.93\scriptsize{(0.01)} & 0.96\scriptsize{(0.01) }& 0.15\scriptsize{(0.03) }&0.15\scriptsize{(0.01)}& 0.14\scriptsize{(0.03) }&  0.14\scriptsize{(0.03) }&    \textbf{0.97\scriptsize{(0.01) }}    \\ 
\cmidrule{2-11}
                        & \textit{CR} $\downarrow$    &\textbf{0.01\scriptsize{(0.03) }}& 0.03\scriptsize{(0.03) } &  \textbf{0.01\scriptsize{(0.03) } } & \textbf{0.01\scriptsize{(0.01) }}& 0.03\scriptsize{(0.02) }& 0.05\scriptsize{(0.01) }& 0.04\scriptsize{(0.02) } & 0.08\scriptsize{(0.02) }&    \textbf{0.01\scriptsize{(0.01) }}       \\ 
\bottomrule
\end{tabular}}}
\caption{Performance of {\name}, planning-based baselines and RL-based baselines with $N=5,20$ agents in OmniDrones.\looseness=-1}}
\label{tab: omnidrones}
\vspace{-5mm}
\end{table*}

%% file: tables/mpe_varying.tex
\begin{table}
\centering
\footnotesize
\vspace{2mm}
{\color{black}
\scalebox{0.96}{
\setlength\tabcolsep{1.5mm}{\begin{tabular}{crccc|c} 
\toprule
Agents                                                                                        & Metrics                  &(H)ORCA  &(H)RRT* &(H)Voronoi  & {\name}  \\ 
\midrule

\multirow{3}{*}{$ 20 \Rightarrow 10 $}& \textit{Steps} $\downarrow$   
                 &43.64\scriptsize{(3.86)}&40.68\scriptsize{(3.37)} & 40.18\scriptsize{(3.41)} & \textbf{22.38\scriptsize{(3.04)}} \\ 
\cmidrule{2-6}
                     & \textit{SR} $\uparrow$     & 0.89\scriptsize{(0.01) }&0.92\scriptsize{(0.01) } &   0.93\scriptsize{(0.01) }&    \textbf{1.00\scriptsize{(0.00) }}    \\ 
\cmidrule{2-6}
                     & \textit{CR} $\downarrow$     & 0.18\scriptsize{(0.02)}& 0.72\scriptsize{(0.14)}& 0.86\scriptsize{(0.13)}  &   \textbf{0.16\scriptsize{(0.03)}} \\ 
\midrule
\multirow{3}{*}{$ 50 \Rightarrow 20 $} & \textit{Steps} $\downarrow$   & $\backslash$ & 89.18\scriptsize{(4.88)} &
88.73\scriptsize{(4.27)}&
\textbf{46.49\scriptsize{(4.49)}}\\ 
\cmidrule{2-6}
                        & \textit{SR} $\uparrow$    & 0.86\scriptsize{(0.01)}&
                       0.91\scriptsize{(0.01} &  0.91\scriptsize{(0.02)}& \textbf{1.00\scriptsize{(0.00)} }\\ 
\cmidrule{2-6}
                     & \textit{CR} $\downarrow$     & \textbf{0.15\scriptsize{(0.03)}}& 0.56\scriptsize{(0.05)}& 0.67\scriptsize{(0.03)}  &   \textbf{0.15\scriptsize{(0.01)}}    \\ 

\bottomrule
\end{tabular}}}}
\caption{Performance of {\name} and planning-based baselines with a varying team size in MPE.}
\vspace{-2mm}
\label{tab: mpe_varying}
\end{table}

%% file: tables/omnidrones_varying.tex
\begin{table}
\centering
\vspace{2mm}
\footnotesize
{\color{black}
\scalebox{0.93}{
\setlength\tabcolsep{1.5mm}{\begin{tabular}{crccc|c} 
\toprule
Agents                                                                                        & Metrics                  &(H)ORCA  &(H)RRT* &(H)Voronoi  & {\name}  \\ 
\midrule

\multirow{3}{*}{$ 20 \Rightarrow 10 $}& \textit{Steps} $\downarrow$   
                  &$\backslash$&$\backslash$ &490.35\scriptsize{(15.63)} & \textbf{426.73\scriptsize{(18.50)}} \\ 
\cmidrule{2-6}
                     & \textit{SR} $\uparrow$    & 0.68\scriptsize{(0.04)  }&0.61\scriptsize{(0.04)  }&   0.85\scriptsize{(0.03) }&    \textbf{0.94\scriptsize{(0.01)}}    \\ 
\cmidrule{2-6}
                     & \textit{CR} $\downarrow$     &\textbf{0.00\scriptsize{(0.01)}} & 0.01\scriptsize{(0.01)}&  0.01\scriptsize{(0.01)}&  0.01\scriptsize{(0.01)}   \\ 
\bottomrule
\end{tabular}}}
\caption{Performance of {\name} and planning-based baselines with a varying team size in OmniDrones.}}
\label{tab:omnidrones_varying}
\vspace{-5mm}
\end{table}

%% file: tables/time_table.tex
\begin{table}
\centering
\footnotesize

\vspace{2mm}
\scalebox{0.82}{
\setlength\tabcolsep{1.0mm}{\begin{tabular}
{c|ccc|ccccc} 
\toprule
Methods & ORCA     & RRT*            & Voronoi        &MAPPO &DARL1N &HTAMP& MAGE-X & {\name}   \\ 
\midrule
Time(ms)    & 39\scriptsize{(6)}  & 456\scriptsize{(10)} & 44\scriptsize{(5)} &\textbf{13\scriptsize{(4)}} & 16\scriptsize{(3)}&14\scriptsize{(3)} &20\scriptsize{(5)} & 14\scriptsize{(2)} \\

\bottomrule
\end{tabular}}}
\caption{Average performance on the per-step time.}
\label{tab: time_table}
\vspace{-6mm}
\end{table}

%% file: tables/ab.tex
\begin{table}
\vspace{2mm}
\centering
\footnotesize

\scalebox{0.87}{
{\color{black}
\setlength\tabcolsep{1.5mm}{\begin{tabular}{crccc|c} 
\toprule
Agents                                                                                        & Metrics                & MASP w. RG &GM w.o. Graph&CAE w.o. Graph& {\name}  \\ 
\midrule

\multirow{3}{*}{N =20}& \textit{Steps} $\downarrow$   
                  &35.91\scriptsize{(2.73)}&$\backslash$ & 28.25\scriptsize{(3.21)} & \textbf{24.41\scriptsize{(2.66)}} \\ 
\cmidrule{2-6}
                     & \textit{SR} $\uparrow$    &\textbf{1.00\scriptsize{(0.00) }}&0.04\scriptsize{(0.01) } &   \textbf{1.00\scriptsize{(0.00) }}&    \textbf{1.00\scriptsize{(0.00) }}    \\ 
\cmidrule{2-6}
                     & \textit{CR} $\downarrow$    &0.25\scriptsize{(0.02) }&3.32\scriptsize{(0.02)} &  0.17\scriptsize{(0.01) } &   \textbf{0.14\scriptsize{(0.01) }  }   \\

\bottomrule
\end{tabular}}}}
\caption{Performance of {\name} and RL variants with $N=20$ agents in MPE. \looseness=-1}
\label{tab: ab}
\vspace{0mm}
\end{table}

%% file: tables/groupsize.tex
\begin{table}
\centering
\footnotesize

\scalebox{1.00}{
{\color{black}
\setlength\tabcolsep{1.5mm}{\begin{tabular}{crccc} 
\toprule
Agents                                                                                        & Metrics                & 2 & 3 & 5  \\ 
\midrule

\multirow{3}{*}{N =20}&    \textit{Steps} $\downarrow$  
                  &28.85\scriptsize{(2.12)} &24.41\scriptsize{(2.66)}  & \textbf{23.78\scriptsize{(2.35)}}    \\ 
\cmidrule{2-5}
                      &   \textit{SR} $\uparrow$  & \textbf{1.00\scriptsize{(0.00) }}  &\textbf{1.00\scriptsize{(0.00) }} & \textbf{1.00\scriptsize{(0.00) }}    \\ 
\cmidrule{2-5}
                      &   \textit{CR} $\downarrow$  &  0.15\scriptsize{(0.01)}  &\textbf{0.14\scriptsize{(0.01)} } &\textbf{0.14\scriptsize{(0.01)} }     \\ 

\bottomrule
\end{tabular}}}}
\caption{Performance of Group Size = 2, 3, and 5.\looseness=-1}
\label{tab: groupsize}
\vspace{-4mm
}
\end{table}

%% file: tables/globalsteps.tex
\begin{table}
\centering
\footnotesize

\scalebox{1.00}{
{\color{black}
\setlength\tabcolsep{1.5mm}{\begin{tabular}{crccc} 
\toprule
Agents                                                                                        & Metrics                & 1 & 3 & 10  \\ 
\midrule

\multirow{3}{*}{N =20}& \textit{Steps} $\downarrow$   
                  & \textbf{24.23\scriptsize{(2.11)}} & 24.41\scriptsize{(2.66)}  & 29.59\scriptsize{(2.43)}    \\ 
\cmidrule{2-5}
                     & \textit{SR} $\uparrow$    & \textbf{1.00\scriptsize{(0.00)}}  & \textbf{1.00\scriptsize{(0.00)}} &\textbf{1.00\scriptsize{(0.00)}}       \\ 
\cmidrule{2-5}
                     & \textit{CR} $\downarrow$    &  \textbf{0.12\scriptsize{(0.01)}}   & 0.14\scriptsize{(0.01)} &    0.16\scriptsize{(0.01)}     \\

\bottomrule
\end{tabular}}}}
\caption{Performance of Global Steps = 1, 3, and 10.\looseness=-1}
\label{tab: globalsteps}
\vspace{-6mm
}
\end{table}

%% file: 6_con.tex
\section{Conclusion and Future Work}
{\color{black}We propose a decentralized goal-conditioned hierarchical planner, the \emph{Multi-Agent Scalable Graph-based Planner}, to improve data efficiency and cooperation in navigation tasks with a substantial number of agents. The high-level policy, the Goal Matcher, leverages a Self-Encoder and a Cross-Encoder to allocate target goals to agents at each global step. The low-level policy, the Coordinated Action Executor, develops the Group Information Fusion to navigate agents toward their designated goals while ensuring effective cooperation.
Thorough experiments demonstrate that {\name} achieves higher task and execution efficiency than planning-based baselines and RL competitors in MPE and Omnidrones with large and varying numbers of agents.}

%% file: root.bbl
\begin{thebibliography}{10}
\providecommand{\url}[1]{#1}
\csname url@rmstyle\endcsname
\providecommand{\newblock}{\relax}
\providecommand{\bibinfo}[2]{#2}
\providecommand\BIBentrySTDinterwordspacing{\spaceskip=0pt\relax}
\providecommand\BIBentryALTinterwordstretchfactor{4}
\providecommand\BIBentryALTinterwordspacing{\spaceskip=\fontdimen2\font plus
\BIBentryALTinterwordstretchfactor\fontdimen3\font minus \fontdimen4\font\relax}
\providecommand\BIBforeignlanguage[2]{{%
\expandafter\ifx\csname l@#1\endcsname\relax
\typeout{** WARNING: IEEEtran.bst: No hyphenation pattern has been}%
\typeout{** loaded for the language `#1'. Using the pattern for}%
\typeout{** the default language instead.}%
\else
\language=\csname l@#1\endcsname
\fi
#2}}

\bibitem{autonomousdriving}
G.~Bresson, Z.~Alsayed, L.~Yu, \emph{et~al.}, ``Simultaneous localization and mapping: A survey of current trends in autonomous driving,'' \emph{IEEE Transactions on Intelligent Vehicles}, vol.~2, no.~3, pp. 194--220, 2017.

\bibitem{autonomousdriving2}
S.~Grigorescu, B.~Trasnea, T.~Cocias, and G.~Macesanu, ``A survey of deep learning techniques for autonomous driving,'' \emph{Journal of Field Robotics}, vol.~37, no.~3, pp. 362--386, 2020.

\bibitem{logistics}
S.~Liu and L.~Hu, ``Application of beidou navigation satellite system in logistics and transportation,'' in \emph{Logistics: The Emerging Frontiers of Transportation and Development in China}, 2009, pp. 1789--1794.

\bibitem{logistics2}
K.~Gao, J.~Xin, H.~Cheng, D.~Liu, and J.~Li, ``Multi-mobile robot autonomous navigation system for intelligent logistics,'' in \emph{2018 Chinese Automation Congress (CAC)}.\hskip 1em plus 0.5em minus 0.4em\relax IEEE, 2018, pp. 2603--2609.

\bibitem{rescue}
A.~Kleiner, J.~Prediger, \emph{et~al.}, ``Rfid technology-based exploration and slam for search and rescue,'' in \emph{2006 IEEE/RSJ International Conference on Intelligent Robots and Systems}.\hskip 1em plus 0.5em minus 0.4em\relax IEEE, 2006, pp. 4054--4059.

\bibitem{rescue2}
D.~Calisi, A.~Farinelli, \emph{et~al.}, ``Autonomous navigation and exploration in a rescue environment,'' in \emph{IEEE International Safety, Security and Rescue Rototics, Workshop, 2005.}\hskip 1em plus 0.5em minus 0.4em\relax IEEE, 2005, pp. 54--59.

\bibitem{frontier3}
W.~Burgard, M.~Moors, C.~Stachniss, and F.~E. Schneider, ``Coordinated multi-robot exploration,'' \emph{IEEE Transactions on robotics}, vol.~21, no.~3, pp. 376--386, 2005.

\bibitem{RRT_Star}
S.~Karaman and E.~Frazzoli, ``Sampling-based algorithms for optimal motion planning,'' \emph{The international journal of robotics research}, vol.~30, no.~7, pp. 846--894, 2011.

\bibitem{maans}
C.~Yu, X.~Yang, J.~Gao, H.~Yang, Y.~Wang, and Y.~Wu, ``Learning efficient multi-agent cooperative visual exploration,'' in \emph{European Conference on Computer Vision}.\hskip 1em plus 0.5em minus 0.4em\relax Springer, 2022, pp. 497--515.

\bibitem{mage-x}
X.~Yang, S.~Huang, Y.~Sun, Y.~Yang, C.~Yu, W.-W. Tu, H.~Yang, and Y.~Wang, ``Learning graph-enhanced commander-executor for multi-agent navigation,'' in \emph{Proceedings of the 2023 International Conference on Autonomous Agents and Multiagent Systems}, 2023, pp. 1652--1660.

\bibitem{inferenceMultiNavigation}
L.~Xia, C.~Yu, and Z.~Wu, ``Inference-based hierarchical reinforcement learning for cooperative multi-agent navigation,'' in \emph{2021 IEEE 33rd International Conference on Tools with Artificial Intelligence (ICTAI)}.\hskip 1em plus 0.5em minus 0.4em\relax IEEE, 2021, pp. 57--64.

\bibitem{mappo}
C.~Yu, A.~Velu, E.~Vinitsky, Y.~Wang, A.~Bayen, and Y.~Wu, ``The surprising effectiveness of ppo in cooperative, multi-agent games,'' \emph{arXiv preprint arXiv:2103.01955}, 2021.

\bibitem{mat}
M.~Wen, J.~G. Kuba, R.~Lin, W.~Zhang, Y.~Wen, J.~Wang, and Y.~Yang, ``Multi-agent reinforcement learning is a sequence modeling problem,'' \emph{arXiv preprint arXiv:2205.14953}, 2022.

\bibitem{multiagent-RL}
C.~Wakilpoor, P.~J. Martin, C.~Rebhuhn, and A.~Vu, ``Heterogeneous multi-agent reinforcement learning for unknown environment mapping,'' \emph{arXiv preprint arXiv:2010.02663}, 2020.

\bibitem{liu2021multi}
X.~Liu, D.~Guo, H.~Liu, and F.~Sun, ``Multi-agent embodied visual semantic navigation with scene prior knowledge,'' \emph{arXiv preprint arXiv:2109.09531}, 2021.

\bibitem{mpe}
R.~Lowe, Y.~I. Wu, A.~Tamar, J.~Harb, O.~Pieter~Abbeel, \emph{et~al.}, ``Multi-agent actor-critic for mixed cooperative-competitive environments,'' \emph{Advances in neural information processing systems}, vol.~30, 2017.

\bibitem{drone}
B.~Xu, F.~Gao, C.~Yu, R.~Zhang, Y.~Wu, and Y.~Wang, ``Omnidrones: An efficient and flexible platform for reinforcement learning in drone control,'' \emph{arXiv preprint arXiv:2309.12825}, 2023.

\bibitem{save}
X.~Yang, C.~Yu, J.~Gao, Y.~Wang, and H.~Yang, ``Save: Spatial-attention visual exploration,'' in \emph{2022 IEEE International Conference on Image Processing (ICIP)}.\hskip 1em plus 0.5em minus 0.4em\relax IEEE, 2022, pp. 1356--1360.

\bibitem{li2020graph}
Q.~Li, F.~Gama, A.~Ribeiro, and A.~Prorok, ``Graph neural networks for decentralized multi-robot path planning,'' in \emph{2020 IEEE/RSJ International Conference on Intelligent Robots and Systems (IROS)}.\hskip 1em plus 0.5em minus 0.4em\relax IEEE, 2020, pp. 11\,785--11\,792.

\bibitem{darl1n}
B.~Wang, J.~Xie, and N.~Atanasov, ``Darl1n: Distributed multi-agent reinforcement learning with one-hop neighbors,'' in \emph{2022 IEEE/RSJ International Conference on Intelligent Robots and Systems (IROS)}.\hskip 1em plus 0.5em minus 0.4em\relax IEEE, 2022, pp. 9003--9010.

\bibitem{kuhn1955hungarian}
H.~W. Kuhn, ``The hungarian method for the assignment problem,'' \emph{Naval research logistics quarterly}, vol.~2, no. 1-2, pp. 83--97, 1955.

\bibitem{turpin2014goal}
M.~Turpin, K.~Mohta, N.~Michael, and V.~Kumar, ``Goal assignment and trajectory planning for large teams of interchangeable robots,'' \emph{Autonomous Robots}, vol.~37, no.~4, pp. 401--415, 2014.

\bibitem{ma2017overview}
H.~Ma, S.~Koenig, N.~Ayanian, L.~Cohen, W.~H{\"o}nig, T.~Kumar, T.~Uras, H.~Xu, \emph{et~al.}, ``Overview: Generalizations of multi-agent path finding to real-world scenarios,'' \emph{arXiv preprint arXiv:1702.05515}, 2017.

\bibitem{wagner2011m}
G.~Wagner and H.~Choset, ``M*: A complete multirobot path planning algorithm with performance bounds,'' in \emph{2011 IEEE/RSJ international conference on intelligent robots and systems}.\hskip 1em plus 0.5em minus 0.4em\relax IEEE, 2011, pp. 3260--3267.

\bibitem{ji2022hierarchical}
Y.~Ji, Z.~Li, Y.~Sun, X.~B. Peng, S.~Levine, G.~Berseth, and K.~Sreenath, ``Hierarchical reinforcement learning for precise soccer shooting skills using a quadrupedal robot,'' in \emph{2022 IEEE/RSJ International Conference on Intelligent Robots and Systems (IROS)}.\hskip 1em plus 0.5em minus 0.4em\relax IEEE, 2022, pp. 1479--1486.

\bibitem{nasiriany2022augmenting}
S.~Nasiriany, H.~Liu, and Y.~Zhu, ``Augmenting reinforcement learning with behavior primitives for diverse manipulation tasks,'' in \emph{2022 International Conference on Robotics and Automation (ICRA)}.\hskip 1em plus 0.5em minus 0.4em\relax IEEE, 2022, pp. 7477--7484.

\bibitem{takubo2022hierarchical}
Y.~Takubo, H.~Chen, and K.~Ho, ``Hierarchical reinforcement learning framework for stochastic spaceflight campaign design,'' \emph{Journal of Spacecraft and rockets}, vol.~59, no.~2, pp. 421--433, 2022.

\bibitem{pope2021hierarchical}
A.~P. Pope, J.~S. Ide, D.~Mi{\'c}ovi{\'c}, \emph{et~al.}, ``Hierarchical reinforcement learning for air-to-air combat,'' in \emph{2021 international conference on unmanned aircraft systems (ICUAS)}.\hskip 1em plus 0.5em minus 0.4em\relax IEEE, 2021, pp. 275--284.

\bibitem{hafner2022deep}
D.~Hafner, K.-H. Lee, I.~Fischer, and P.~Abbeel, ``Deep hierarchical planning from pixels,'' \emph{Advances in Neural Information Processing Systems}, vol.~35, pp. 26\,091--26\,104, 2022.

\bibitem{htamp}
A.~A.~R. Newaz and T.~Alam, ``Hierarchical task and motion planning through deep reinforcement learning,'' in \emph{2021 Fifth IEEE International Conference on Robotic Computing (IRC)}.\hskip 1em plus 0.5em minus 0.4em\relax IEEE, 2021, pp. 100--105.

\bibitem{ppo}
J.~Schulman, F.~Wolski, P.~Dhariwal, A.~Radford, \emph{et~al.}, ``Proximal policy optimization algorithms,'' \emph{arXiv preprint arXiv:1707.06347}, 2017.

\bibitem{gerkey2004formal}
B.~P. Gerkey and M.~J. Matari{\'c}, ``A formal analysis and taxonomy of task allocation in multi-robot systems,'' \emph{The International journal of robotics research}, vol.~23, no.~9, pp. 939--954, 2004.

\bibitem{Hungarian}
R.~Burkard, M.~Dell'Amico, and S.~Martello, \emph{Assignment problems: revised reprint}.\hskip 1em plus 0.5em minus 0.4em\relax SIAM, 2012.

\bibitem{attention}
A.~Vaswani, N.~Shazeer, \emph{et~al.}, ``Attention is all you need,'' \emph{Advances in neural information processing systems}, vol.~30, 2017.

\bibitem{orca}
K.~Guo, D.~Wang, T.~Fan, and J.~Pan, ``Vr-orca: Variable responsibility optimal reciprocal collision avoidance,'' \emph{IEEE Robotics and Automation Letters}, vol.~6, no.~3, pp. 4520--4527, 2021.

\bibitem{RRT}
J.~J. Kuffner and S.~M. LaValle, ``Rrt-connect: An efficient approach to single-query path planning,'' in \emph{Proceedings 2000 ICRA. Millennium Conference. IEEE International Conference on Robotics and Automation. Symposia Proceedings (Cat. No. 00CH37065)}, vol.~2.\hskip 1em plus 0.5em minus 0.4em\relax IEEE, 2000, pp. 995--1001.

\bibitem{Voronoi}
J.~Hu, H.~Niu, J.~Carrasco, B.~Lennox, and F.~Arvin, ``Voronoi-based multi-robot autonomous exploration in unknown environments via deep reinforcement learning,'' \emph{IEEE Transactions on Vehicular Technology}, vol.~69, no.~12, pp. 14\,413--14\,423, 2020.

\end{thebibliography}
